\documentclass{naturep}

\usepackage{amssymb}
\usepackage{amsmath}
\usepackage{graphicx}
\usepackage{algorithmicx}
\usepackage{algorithm}
\usepackage[noend]{algpseudocode}
\usepackage{bbm}
\usepackage{lineno}
\usepackage[margin=0.6in]{geometry}
\usepackage{ragged2e}
\usepackage{setspace}
\usepackage{longtable}
\usepackage{hyperref}
\usepackage{anyfontsize}
\usepackage{float}
\usepackage{booktabs}
\usepackage{multirow}
\usepackage{xcolor}
\usepackage{blindtext}
\usepackage{tabularx}
\usepackage{caption}
\usepackage{array}
\usepackage{bbding}
\usepackage{pifont}
\usepackage{fontawesome}
\usepackage{xurl}

\usepackage[normalem]{ulem}

\makeatletter
\let\saved@includegraphics\includegraphics
\AtBeginDocument{\let\includegraphics\saved@includegraphics}

\makeatother

\usepackage{enumitem, kantlipsum}

\title{NeuroClaw Technical Report\\{\large Closed-Loop Agentic AI for Executable and Reproducible Neuroimaging Research}}

\begin{document}
% \linenumbers

\maketitle

\begingroup
\setlength{\parindent}{0pt}
\setlength{\parskip}{2pt}
\vspace{4mm}
\begin{spacing}{1.4}
Cheng Wang$^{1,\dagger}$, Zhibin He$^{2,\dagger}$, Zhihao Peng$^{1,\dagger}$, Shengyuan Liu$^{1}$, Yufan Hu$^{1}$, Carl Yang$^{3}$, Lifang He$^{4}$, Lichao Sun$^{4}$, Xiang Li$^{5,6*}$, Yixuan Yuan$^{1*}$

\begin{affiliations}
 \item Electronic Engineering Department, The Chinese University of Hong Kong, China
 \item School of Electronic Information, Northwest University, China
 \item Department of Computer Science, Emory University, Atlanta, GA, USA
 \item Computer Science and Engineering, Lehigh University, Bethlehem, PA, USA
 \item Department of Radiology, Massachusetts General Hospital, Boston, MA, USA
 \item Kempner Institute for Natural and Artificial Intelligence, Harvard University, Cambridge, MA, USA
\end{affiliations}

{$^\dagger$ These authors contributed equally as co-first authors.}

{*Corresponding authors}
\end{spacing}
\endgroup

\footnotetext[1]{
Xiang Li: \nolinkurl{xli60@mgh.harvard.edu}; 
Yixuan Yuan: \nolinkurl{yxyuan@ee.cuhk.edu.hk}.
}

\vspace{-5mm}
\begin{spacing}{1.0}

\section{Abstract}
Agentic artificial intelligence systems promise to accelerate scientific workflows, but neuroimaging poses unique challenges: heterogeneous modalities (sMRI, fMRI, dMRI, EEG), long multi-stage pipelines, and persistent reproducibility risks. To address this gap, we present NeuroClaw, a domain-specialized multi-agent research assistant for executable and reproducible neuroimaging research. NeuroClaw operates directly on raw neuroimaging data across formats and modalities, grounding decisions in dataset semantics and BIDS metadata so users need not prepare curated inputs or bespoke model code. The platform combines harness engineering with end-to-end environment management, including pinned Python environments, Docker support, automated installers for common neuroimaging tools, and GPU configuration. In practice, this layer emphasizes checkpointing, post-execution verification, structured audit traces, and controlled runtime setup, making toolchains more transparent while improving reproducibility and auditability. A three-tier skill/agent hierarchy separates user-facing interaction, high-level orchestration, and low-level tool skills to decompose complex workflows into safe, reusable units. Alongside the NeuroClaw framework, we introduce NeuroBench, a system-level benchmark for executability, artifact validity, and reproducibility readiness. Across multiple multimodal LLMs, NeuroClaw-enabled runs yield consistent and substantial score improvements compared with direct agent invocation. Project homepage: \url{https://cuhk-aim-group.github.io/NeuroClaw/index.html}.

\end{spacing}

\newpage

\section{Introduction}
% 神经科学研究中，biomarker 的发现费人费力，同时还有严重的复现性问题。引用大量 test-retest 的文章，和评测深度学习方法的文章
Neuroimaging has become a central paradigm for studying brain organization, disease mechanisms, and treatment response across modern neuroscience and clinical research. These workflows generate high-dimensional multimodal data, including structural MRI, functional MRI, diffusion MRI, EEG, MEG, etc., enabling researchers to investigate anatomy, brain activity, connectivity, and microstructure at multiple scales~\cite{ren2025deepprep}. It is these datasets that drive valuable scientific investigation, enabling disease prediction, phenotype analysis, brain network analysis, and other downstream studies~\cite{ren2025deepprep}. However, translating these signals into robust and reproducible neuroscientific findings has remained difficult in practice. Test-retest reliability is often limited, and reproducibility can vary substantially across sessions, sites, scanners, analysis choices, and sample sizes~\cite{brandt2013testretest,chen2007testretest,hougaard2023lack,taxali2021boost,noble2021guide,herting2018testretest,chen2018reproducibility}. Existing studies therefore show both the promise and the fragility of current neuroimaging workflows: deep-learning-enabled systems such as DeepPrep improve the speed and scale of preprocessing~\cite{ren2025deepprep}, whereas reliability studies continue to document instability in task fMRI, resting-state fMRI, and longitudinal designs~\cite{brandt2013testretest,chen2007testretest,hougaard2023lack,herting2018testretest}. Yet these advances remain difficult to operationalize at scale because complex dependencies, fragile runtime environments, and the lack of standardized validation pipelines often prevent promising methods from becoming executable and reproducible research workflows~\cite{nbme,hu2025reprobench}.

% 近期 agentic AI systems 发展迅速，为解决这两个问题带来希望
At the same time, multi-agent systems have advanced rapidly and are beginning to automate increasingly complex scientific workflows. The empowering framework~\cite{gao2024empowering} outlines a biomedical AI scientist paradigm in which collaborative agents integrate models, domain tools, and experimental platforms to support skeptical reasoning and long-horizon discovery. Biomni~\cite{biomni} extends this direction as a general-purpose biomedical agent for tasks such as hypothesis generation, protocol design, clinical reasoning, gene prioritization, drug repurposing, rare disease diagnosis, and microbiome analysis. OriGene~\cite{origene} focuses more narrowly on therapeutic target discovery, using a self-evolving multi-agent system to reason over heterogeneous biomedical evidence and identify mechanistically grounded disease targets. Outside biomedicine, AutoSOTA~\cite{autosota} automates the pipeline from paper understanding to environment repair, experiment execution, and iterative model optimization for state-of-the-art AI systems. EvoScientist~\cite{evoscientist} couples researcher, engineer, and evolution-manager agents with persistent ideation and experimentation memory so that research strategies improve over time rather than remaining fixed. The AI Scientist~\cite{natureauto} further shows that idea generation, coding, experimentation, and paper drafting can be organized into an end-to-end research loop, while Ax-Prover~\cite{axprover} demonstrates that agentic workflows can combine large language models with formal verification tools to solve theorem-proving tasks under strict correctness constraints. These studies suggest that role-specialized agents with shared memory, tool access, and iterative feedback can support long-horizon scientific workflows beyond the reach of single-model prompting or manual scripting.

% 现有 agentic 还完全不足以在复杂的神经科学研究中使用
However, existing agentic systems do not fully meet the needs of neuroscience research. Most emphasize general interaction fluency, broad task completion rates, or hypothesis generation~\cite{triagent,brad,heal,sage,automed,tissuelab,biomni,origene}, but pay limited attention to strict dataset semantics, BIDS-compliant metadata consistency, stage-wise artifact validity, and cross-environment reproducibility~\cite{gorgolewski2016bids,esteban2019fmriprep,cieslak2021qsiprep}. In neuroimaging, even small metadata inconsistencies or environment drift can invalidate downstream analyses, and these errors may become apparent only in later pipeline stages. As a result, current frameworks often produce plausible yet scientifically unusable outputs and do not support an automated experimental discovery loop on neuroscience data from raw data preparation to model feedback and protocol adjustment.

% 我们提出 NeuroClaw，说明特点：可处理原始数据、模型的可操作和可复现、合理的三层技能设计
To address this gap, we present NeuroClaw, a research assistant for executable and reproducible neuroimaging research. Its design centers on three practical advantages. First, NeuroClaw works directly from raw neuroimaging data across formats, modalities, and long multi-stage workflows, so users do not need to manually preprocess inputs or rewrite model-specific data pipelines in advance. Second, NeuroClaw recognizes that neuroimaging experimentation depends not only on model design but also on reliable environment, toolchain, and execution management. It manages Python environments, Docker containers, GPU stacks, and neuroimaging toolchains such as FSL, FreeSurfer, and fMRIPrep\footnote{Official project websites: FSL, \url{https://fsl.fmrib.ox.ac.uk/fsl/fslwiki}; FreeSurfer, \url{https://surfer.nmr.mgh.harvard.edu/}; and fMRIPrep, \url{https://fmriprep.org/}.} so that users do not need to know which tool to invoke or how to configure the runtime across both medical-imaging and computational dependencies. This layer is further constrained by harness engineering, including checkpointed execution, structured verification, and audit logging, to make model execution and experimental results more reproducible. Third, NeuroClaw adopts a hierarchical skill architecture inspired by systems such as Google ADK, but specialized for neuroimaging research: a interface layer for intent handling, a sub-agent layer for orchestration, and a base-execution layer for concrete operations. Together, these components support a self-driven experimental loop in which the agent proposes analysis actions conditioned on dataset semantics, executes tool-grounded steps, verifies artifacts and metrics, and revises subsequent actions across a complete neuroscience discovery workflow. The framework is intended to support neuroscience experimentation rather than pipeline completion alone.

% 配合 framework 设计，我们还提出 benchmark
We also introduce NeuroBench, a benchmark with standardized specifications that assesses executability, output validity, and reproducibility readiness under realistic conditions~\cite{jiang2025medagentbench,zhu2025medagentboard,hu2025reprobench}. As illustrated in Figure~\ref{fig:main}, NeuroClaw couples raw-data-aware workflow orchestration, hierarchical skills, environment-aware execution, and standardized evaluation into a unified loop. NeuroBench provides a standardized way to test whether an agent can produce specification-compliant artifacts across major modalities and representative datasets, and whether it can sustain experiment-loop behavior under realistic constraints.

\begin{figure*}[t]
\centering
\includegraphics[width=0.98\linewidth]{fig/framework.pdf}
\caption{NeuroClaw system framework for executable and reproducible agentic neuroimaging research. (a) Multimodal raw data processing pipeline that converts sMRI, fMRI, DWI, and other inputs into analysis-ready outputs through BIDS organization, quality control, registration, and feature construction. (b) Model executability and reproducibility harness, including single-command execution, pinned dependencies, Docker support, provenance logging, and cross-environment verification. (c) Hierarchical skill architecture with three layers (interface, subagent, and base) that decomposes tasks into tool, modality, dataset, and model skills. (d) NeuroBench evaluation showing benchmark scores and performance gains of various multimodal large language models when using NeuroClaw skills compared with vanilla baselines. The overall design enables an iterative experimental loop of planning, execution, verification, and refinement rather than one-pass completion.}
\label{fig:main}
\end{figure*}

\section{Related Work}
% agent 工作和 autoresearch 工作
\subsection{Agentic Systems for Biomedical and Scientific Discovery.}
A series of recent agentic frameworks have shown strong capabilities in biomarker discovery, medical diagnosis, and automated scientific research. TriAgent~\cite{triagent} employs a dedicated agent for novel biomarker discovery from clinical data and a second agent for literature retrieval, validation, and novelty assessment. BRAD~\cite{brad} delivers an end-to-end pipeline for automatic biomarker discovery and enrichment with automated report generation. HEAL-KGGen~\cite{heal} introduces a hierarchical agentic framework augmented by a constructed medical knowledge graph for genetic biomarker-based diagnosis. SAGE~\cite{sage} proposes an agentic system for interpretable and clinically translatable computational pathology biomarker discovery through direct programming. Auto-MedCalc~\cite{automed} leverages AI agents for automated biomarker discovery and risk score generation from gene datasets. A co-evolving agentic system~\cite{tissuelab} targets pathology, radiology, and spatial omics with continuous expert feedback. Biomni~\cite{biomni} is a general-purpose biomedical agent supporting hypothesis generation, protocol design, clinical reasoning, gene prioritization, drug repurposing, rare disease diagnosis, and microbiome analysis. OriGene~\cite{origene} functions as a self-evolving virtual disease biologist for automated target discovery. Additionally, the NBME copilot~\cite{nbme} focuses on enhancing the reliability of large language models as data-science programming assistants for biomedical research.

Beyond biomedicine, a parallel line of autoresearch systems aims at end-to-end scientific automation. AutoSOTA~\cite{autosota} generates improved state-of-the-art AI models directly from published papers with an emphasis on harness engineering. The self-evolving recommendation system~\cite{selfrec} uses LLM agents for autonomous model optimization, throughput improvement, and real-world metric validation. EvoScientist~\cite{evoscientist} explores multi-agent evolving AI scientists for end-to-end discovery. Complementary high-impact works include end-to-end automation of AI research~\cite{natureauto}, multi-agent discovery of mathematical concepts~\cite{mathmulti}, analyses of artificial intelligence and the structure of mathematics~\cite{aimath}, and Ax-Prover~\cite{axprover}, an agentic Lean proving system that combines LLMs with MCP-based verifiers. While these systems establish important precedents for tool orchestration, self-evolution, and autonomous discovery, they generally prioritize broad research assistance, hypothesis generation, or model-level optimization. They offer limited support for the strict dataset semantics, BIDS-compliant metadata consistency, stage-wise artifact validity, and heterogeneous environment control that are essential for reliable neuroimaging pipelines.

\subsection{OpenClaw Ecosystems and Autoresearch.}
% 各种 claw 工作和 benchmark 工作
Representative frameworks in the OpenClaw ecosystem include OmicClaw for omics data analysis, EthoClaw for automated neuroethology workflows, and Dr.~Claw for general AI research from idea to paper. Together with extensions such as AutoResearchClaw and hospital-oriented OpenClaw variants, these systems show how modular skill libraries and reusable workflows can support long-horizon scientific automation. A related but broader autoresearch line pursues end-to-end scientific automation, including AutoResearch for fully autonomous experimentation on single GPUs, AutoSOTA~\cite{autosota}, and EvoScientist~\cite{evoscientist}. These systems extend the same general paradigm to paper understanding, environment repair, experiment execution, persistent memory, and iterative refinement. Related biomedical agent frameworks reinforce this direction. The biomedical AI scientist perspective in Cell~\cite{gao2024empowering}, coordinated healthcare-agent architectures in Nature Biomedical Engineering~\cite{moritz2025coordinated}, and foundational healthcare-agent designs in Cell Reports Medicine~\cite{liu2025foundational} all emphasize role specialization, structured tool use, and explicit coordination. Clinical systems such as autonomous oncology decision agents~\cite{ferber2025autonomous} and EvoMDT~\cite{liu2026evomdt} further show that deployment depends on traceable execution and controllable collaboration. Collectively, these OpenClaw and autoresearch efforts show the promise of reusable skills and multi-agent orchestration, while also exposing the need for stronger guarantees of executability, auditability, and robustness under real scientific toolchains.

Recent benchmarks and evaluation frameworks show that this reliability problem remains open. ClawBench~\cite{clawbench} is particularly informative because it evaluates 153 everyday online tasks across 144 live platforms rather than static sandbox environments, revealing that even frontier models achieve only modest success rates (for example, Claude-Sonnet-4.6~\cite{anthropic2025claude} reaches 33.3\%). REPRO-Bench~\cite{hu2025reprobench} makes a complementary point from the reproducibility perspective: beyond producing plausible outputs, agentic systems must assess whether workflows can actually be rerun, verified, and trusted. Studies of executable discovery systems further reinforce this distinction between apparent competence and dependable scientific execution. Virtual Lab-style work~\cite{swanson2025virtuallab} demonstrates that AI agents can participate in hypothesis-driven discovery loops with experimental validation, while agentic bioinformatics pipelines~\cite{zhou2025agenticbio} show how workflow automation can accelerate biomedical analysis at scale. Yet these results also underline the central benchmarking gap: current evaluations rarely stress heterogeneous environments, staged artifact validation, and cross-run reproducibility with the granularity required by neuroimaging workflows. Collectively, these benchmarks and executable-agent studies motivate NeuroBench as a more domain-grounded evaluation setting focused on long-horizon executability, reproducibility readiness, and robustness to the practical failures that arise when agents operate directly on scientific data and toolchains.

\section{NeuroClaw}
NeuroClaw is a research assistant for neuroscience designed for reliable execution and reproducible neuroimaging workflows. It aims not only to complete individual tasks but also to support a complete neuroscience discovery workflow with iterative scientific experimentation and auditability. The system is designed to start from raw data and tool-accessible model assets, without requiring users to prepare preprocessed inputs or customized model code in advance. The platform emphasizes three design components: raw-data-aware orchestration for complex multimodal neuroimaging data, environment-aware experimentation with harness engineering for executable and reproducible runs, and a hierarchical skill architecture that balances modularity and control.

\subsection{Skill Hierarchy.}
% 三层结构，合理的拆解任务和互相写作
NeuroClaw organizes its capabilities into a three-layer skill hierarchy that separates concerns and scales with task complexity. This design is conceptually aligned with hierarchical agent-development patterns such as Google ADK, but is specialized for neuroimaging research requirements. The core design principle is to decompose long neuroimaging workflows into shorter skills with clear interfaces, so that the agent retrieves only the operations needed for the current stage rather than reasoning over a monolithic prompt-sized procedure. The \textbf{base layer} provides fine-grained atomic operations such as file conversion, DICOM-to-NIfTI transformation, metadata validation, and single-command execution. These operations serve as safe, reusable building blocks. The \textbf{subagent layer} implements domain-specific orchestration through four skill types: \emph{tool} skills that wrap external software such as FSL, FreeSurfer, and fMRIPrep; \emph{modality} skills that encapsulate preprocessing pipelines for fMRI, sMRI, and DWI; \emph{model} skills that handle phenotype prediction and statistical analysis; and \emph{dataset} skills that coordinate end-to-end multimodal workflows for cohorts such as ADNI, HCP Young Adult, and UK Biobank. The \textbf{interface layer} manages high-level user interaction by interpreting intent, routing tasks to appropriate subagents, and maintaining workflow continuity without exposing low-level implementation details. This hierarchy improves skill selection efficiency, reduces execution drift, enables independent maintenance of evolving toolchains, and keeps iterative updates localized and controllable. It also avoids overly long single skills that are harder for models to parse, more likely to dilute retrieval focus, and more expensive in token usage.

All skill dependencies are structured as a Directed Acyclic Graph (DAG). This design enforces one-way information flow, prevents cyclic dependencies, and supports efficient skill invocation through minimal traversal over only the relevant predecessors and successors for a given task. In practice, the DAG makes skill relations explicit, so the agent can compose multi-step workflows by following dependency structure instead of repeatedly reinterpreting a large undifferentiated skill description. Combined with the three-layer hierarchy, this organization decomposes complex neuroimaging workflows into compact, well-defined units, improving execution reliability and decision accuracy in long-horizon tasks while reducing model confusion and unnecessary token consumption.

\subsection{Raw-Data Compatibility.}
% 覆盖神经影像的各种数据模态，不需要人为预处理和格式化
A core feature of NeuroClaw is its ability to work directly from raw data. Rather than assuming preprocessed inputs or routing solely according to tool names or free-form instructions, the system conditions planning and skill selection on explicit data context, including data organization, BIDS compliance, available modalities, preprocessing stage, and metadata conventions. For example, when a user specifies the HCP Young Adult dataset, the corresponding `hcp-skill` automatically constructs a multimodal processing plan spanning structural MRI, functional MRI, and diffusion MRI, then delegates each stage to the appropriate modality skill. This approach reduces ambiguity in heterogeneous computing environments, where the validity of a command often depends on upstream data conventions. By grounding decisions in data semantics from the outset and supporting workflows that begin with raw inputs, NeuroClaw reduces cascading errors, helps intermediate artifacts satisfy downstream requirements, and maintains scientific coherence across iterative refinements. Parallel efforts in computational neuroscience further highlight the importance of scalable infrastructure for reproducible analysis, such as high-performance frameworks for neural population dynamics modeling with optimized hyperparameter tuning across diverse computing environments~\cite{patel2023neuralpop}.

NeuroClaw supports end-to-end workflows across major neuroimaging modalities. For structural MRI (sMRI), it handles DICOM-to-NIfTI conversion, BIDS organization, skull stripping, tissue segmentation, cortical surface reconstruction, and ROI-based statistical extraction. For functional MRI (fMRI), it supports preprocessing and denoising pipelines, confound regression, ROI time-series extraction, connectivity analysis, and first- to group-level GLM reporting. For diffusion MRI (dMRI), the framework manages preprocessing, tensor fitting, microstructural metric computation, tractography, and connectome construction. For EEG, it provides standard preprocessing, artifact removal, spectral feature extraction, and feature table generation. This unified harness-driven approach enables consistent workflows across modalities and supports iterative hypothesis testing through shared data-model interfaces.

For dataset-specific workflows, NeuroClaw provides tailored orchestration for ADNI, HCP, and UK Biobank cohorts. In ADNI-oriented processing, it guides users through data staging, BIDS-compliant reorganization, modality-aware preprocessing, and specification-aligned generation of ROI features and quality-control summaries, while correctly interpreting cohort-specific metadata conventions. In HCP-oriented processing, it coordinates long-horizon multimodal pipelines, manages stage dependencies, and preserves consistent intermediate artifacts for downstream analysis and reproducibility verification. For UK Biobank, NeuroClaw supports brain-focused downstream analysis on already available local UKB-derived tables and feature matrices, including neurological outcome modeling, cognitive phenotype analysis, brain MRI derived phenotype association, and predictive modeling. Across all three cohorts, the framework emphasizes auditable execution traces, validation checkpoints, and standardized outputs, allowing users to conduct repeated experimental cycles from raw data or analysis-ready tables to model-ready results with substantially reduced manual effort.

\subsection{Reproducibility and Harness Engineering.}
% 如何保证可操作和可复现，包括环境管理，harness 约束
NeuroClaw promotes reproducibility through a concrete harness layer that couples runtime control with post-hoc validation. At the session level, it supports lightweight checkpointing and resumability through timestamped state snapshots. At the workflow level, experiment execution follows an explicit decomposition-initialization-execution-verification protocol with environment manifests, structured logging, per-stage checkpoints, and final audit reports. Across major tool and modality skills, NeuroClaw further standardizes post-execution verification through expected-artifact checks, missing-file detection, NaN/Inf screening, quality-control validation, and structured JSONL audit logs. Complementary runtime components add drift logging, checksum-based artifact validation, and container isolation or image-integrity checks where needed. Together with installer-managed environments for major neuroimaging toolchains, these mechanisms make iterative workflows easier to rerun, inspect, and compare across executions.

\begin{figure*}[t]
\centering
\includegraphics[width=0.95\linewidth]{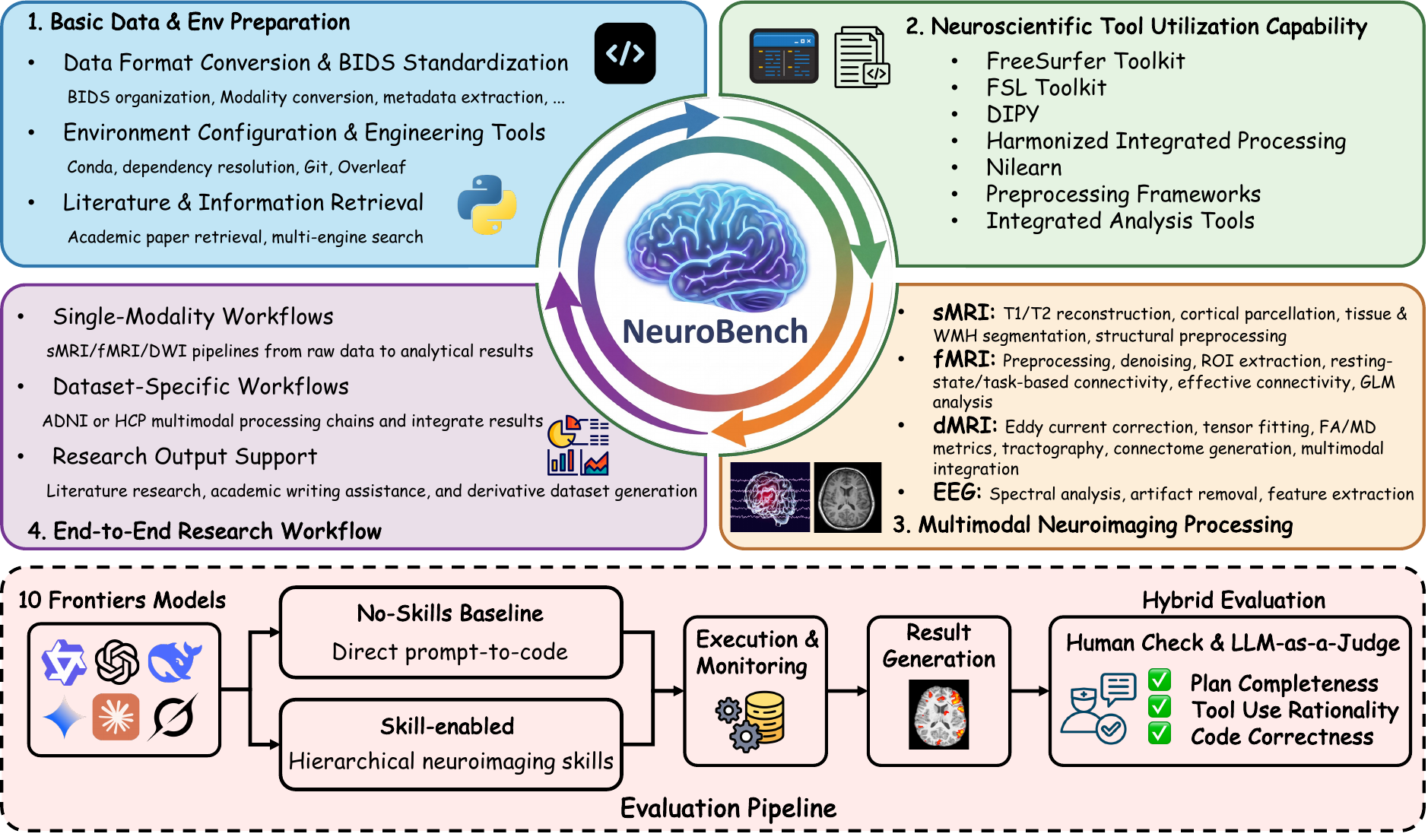}
\caption{Overview of NeuroBench. The benchmark is organized into four modules (top): basic data and environment preparation, neuroscientific tool utilization, multimodal neuroimaging processing, and research-loop execution. The bottom panel illustrates the evaluation pipeline, in which frontier multimodal large language models are assessed under both baseline (no-skills) and skill-enabled (with NeuroClaw) settings. The pipeline includes execution monitoring, standardized result generation, and hybrid evaluation combining human experts and MLLM judges across planning completeness, tool usage reasonableness, and command or code correctness.}
\label{fig:benchmark}
\end{figure*}

\section{NeuroBench Benchmark}
NeuroBench is a system-level benchmark designed to evaluate whether an agent can complete realistic neuroimaging research tasks under practical constraints. As shown in Figure~\ref{fig:benchmark}, it organizes evaluation into four modules: basic data and environment preparation, neuroscientific tool utilization, multimodal neuroimaging processing, and research-loop execution. The benchmark focuses on closed-loop execution capability, spanning task understanding, planning, command and code generation, artifact validation, and iteration readiness, rather than isolated tool calls or static question answering.
\subsection{Benchmark Design.}

% 100 个任务覆盖范围
All 100 tasks (T001-T100) are manually designed by experienced neuroimaging researchers, ensuring scientific validity and practical relevance. Each task is paired with a hand-crafted evaluation specification that explicitly defines input assumptions (data organization, BIDS compliance status, and metadata requirements), processing objectives, expected output artifacts (processed images, ROI tables, connectomes, and QC reports), naming conventions, and verifiable checkpoints. This specification-driven design clarifies the criteria for successful completion and enables reliable automated verification of output conformance. Tasks span major modalities, including structural MRI, functional MRI, diffusion MRI, and EEG, and cover representative open-science datasets such as ADNI and HCP Young Adult, with explicit consideration of multi-site heterogeneity, scanner harmonization, and clinical preprocessing standards.

% 任务和 skill 结构的对应
Tasks are organized hierarchically into workflow families: basic utilities (BIDS validation, format conversion, and metadata extraction), core pipelines (ANTs alignment, FSL preprocessing, FreeSurfer cortical reconstruction, fMRIPrep, and QSIPrep), advanced analysis (seed-based connectivity, ICA-FIX denoising, first-level task fMRI GLM, and ROI-wise statistics), and multimodal integration (structural-to-functional registration, diffusion-to-structural alignment, and multi-modal neuromarker computation). This structure comprehensively tests a wide range of agent capabilities, including fine-grained command composition, complex parameter selection, modality-specific domain knowledge, long-horizon planning across preprocessing stages, and reproducible output validation. The human-curated design ensures that every task reflects genuine neuroscience research workflows instead of synthetic or overly simplified scenarios.

\subsection{Scoring Mechanism.}
% 评测方式的介绍
NeuroBench adopts a three-layer scoring structure consisting of LLM-based quality scoring, runtime and efficiency statistics, and paired comparison between \textit{with-skills} and \textit{no-skills} settings. To ensure fairness, scoring is conducted exclusively on the common task subset completed by all models, with repeated runs aligned to the minimum common repetition count (typically five). For each run, a fixed GPT-5.4 judge evaluates the generated report on three dimensions using a 1-10 scale: planning completeness ($P$), tool and skill usage reasonableness ($R$), and command or code correctness ($C$). The weighted score is calculated as $S_{10} = 0.30 \cdot P + 0.40 \cdot R + 0.30 \cdot C$ and converted to a percentage scale via $S_{100} = 10 \cdot S_{10}$.

Skill call count, runtime, and token consumption are recorded as auxiliary efficiency metrics but are not incorporated into the primary quality score. For each model-task pair, we compute the mean and variance of the weighted score across repetitions, together with average skill calls, token usage, and elapsed time. Model-level performance is obtained by averaging across all tasks. Final ranking follows a lexicographic order: (1) higher average quality score (descending), (2) fewer average skill calls (ascending), (3) fewer average tokens (ascending), and (4) shorter runtime (ascending), thereby prioritizing solution quality before operational efficiency.

For the \textit{with-skills} versus \textit{no-skills} comparison on the same base model, we report both absolute gain $A_{\text{abs}} = S_{\text{with}} - S_{\text{no}}$ and normalized gain

\begin{equation}
    g = 
    \begin{cases} 
    \frac{\Delta}{100 - S_{\text{no}}} & \text{if } \Delta \ge 0, \\
    \frac{\Delta}{S_{\text{no}}} & \text{if } \Delta < 0.
    \end{cases}
    \label{eq:gain}
\end{equation}

\noindent where $S_{\text{with}}$ and $S_{\text{no}}$ denote the model's average score under the \textit{with-skills} and \textit{no-skills} settings, respectively, $\Delta = S_{\text{with}} - S_{\text{no}}$, and $g$ is clipped to the range $[-1, 1]$.

\subsection{Results.}
\begin{figure*}[t]
\centering
\includegraphics[width=1.\linewidth]{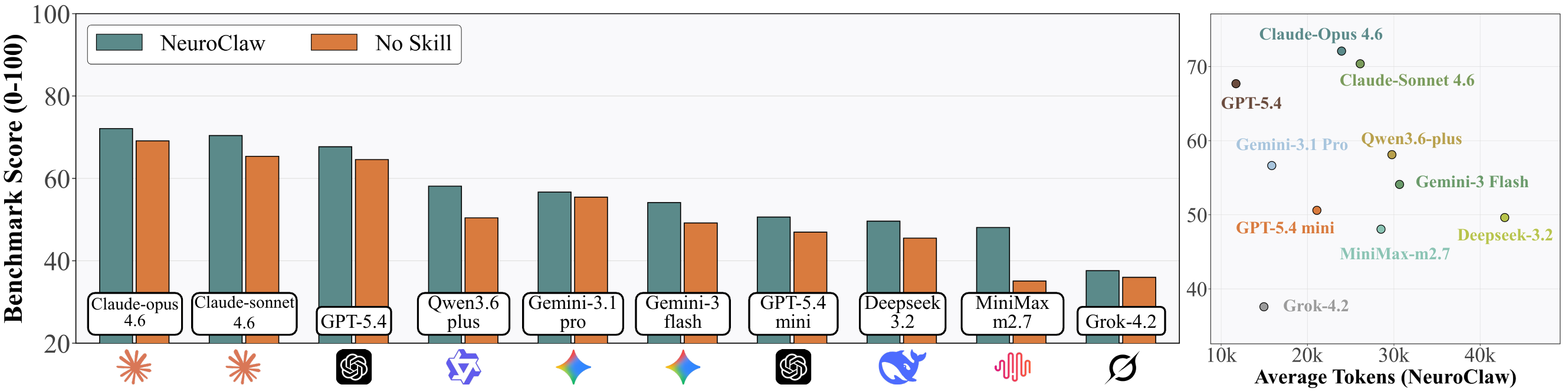}
\caption{Model performance on NeuroBench under \textit{with-skills} and \textit{no-skills} settings. Here, \textit{with-skills} denotes the corresponding base model running within the NeuroClaw framework. (Left) Overall benchmark scores for each base model under the two settings, shown on a 0--100 scale. (Right) Trade-off between overall performance and efficiency under the \textit{with-skills} setting, where each base model's average score is plotted against its average token consumption; point labels indicate the corresponding multimodal large language model.}
\label{fig:three-dim-scores}
\end{figure*}

Figure~\ref{fig:three-dim-scores} compares each base model under the \textit{with-skills} and \textit{no-skills} settings, where \textit{with-skills} denotes the same base model running within the NeuroClaw framework. The left panel summarizes overall benchmark scores, and the right panel shows the trade-off between score and token consumption under the \textit{with-skills} setting. NeuroBench still evaluates each run on three dimensions, namely planning completeness ($P$), tool/skill usage reasonableness ($R$), and command/code correctness ($C$), and we discuss these component scores below together with the aggregate gains reported in Table~\ref{tab:neurobench-skillgain}.

\begin{table}[!t]
\centering
\caption{Performance gain from skill usage on NeuroBench. $A_{\text{abs}}$ denotes absolute score improvement over the \textit{no-skills} baseline, and $g$ is the normalized gain.}
\label{tab:neurobench-skillgain}
\begin{tabular}{l c c c c}
\toprule
\textbf{Base Model} & \textbf{With Skills (\%)} & \textbf{No Skills (\%)} & \textbf{$A_{\text{abs}}$ (\%)} & \textbf{$g$} \\
\midrule
Claude-Opus-4.6~\cite{anthropic2025claude} & 72.10 & 69.12 & 2.98 & 0.0965 \\
Claude-Sonnet-4.6~\cite{anthropic2025claude} & 70.39 & 65.37 & 5.02 & 0.1450 \\
DeepSeek-3.2~\cite{deepseek2025v32} & 49.63 & 45.49 & 4.14 & 0.0759 \\
Gemini-3-Flash~\cite{google2025gemini} & 54.10 & 49.15 & 4.95 & 0.0973 \\
Gemini-3.1-Pro~\cite{google2025gemini} & 56.65 & 55.43 & 1.22 & 0.0274 \\
GPT-5.4~\cite{openai2025gpt5} & 67.69 & 64.57 & 3.12 & 0.0881 \\
GPT-5.4-mini~\cite{openai2025gpt5} & 50.61 & 46.94 & 3.67 & 0.0692 \\
Grok-4.2~\cite{xai2025grok4} & 37.59 & 35.97 & 1.62 & 0.0253 \\
MiniMax-M2.7~\cite{minimax2026m2} & 48.07 & 35.10 & 12.97 & 0.1998 \\
Qwen3-plus~\cite{yang2025qwen3} & 58.12 & 50.39 & 7.73 & 0.1558 \\
\bottomrule
\end{tabular}
\end{table}

% 这一段分析性能结果
Figure~\ref{fig:three-dim-scores} shows clear overall performance differences among models under the \textit{with-skills} and \textit{no-skills} settings. Claude-Opus-4.6~\cite{anthropic2025claude} achieves the highest overall score under the NeuroClaw setting (72.10\%), followed by Claude-Sonnet-4.6~\cite{anthropic2025claude} (70.39\%) and GPT-5.4~\cite{openai2025gpt5} (67.69\%). Table~\ref{tab:neurobench-skillgain} shows that all ten base models improve when run within the NeuroClaw framework, with an average absolute gain of 4.74 points. Larger gains are observed for Qwen3-plus~\cite{yang2025qwen3} (7.73 points) and MiniMax-M2.7~\cite{minimax2026m2} (12.97 points), whereas GPT-5.4~\cite{openai2025gpt5} shows a gain of 3.12 points with a normalized gain of $g = 0.0881$. At the same time, the evaluation remains grounded in the three component scores $P$, $R$, and $C$, which help explain where the aggregate improvements come from even though they are no longer plotted directly in Figure~\ref{fig:three-dim-scores}.

% 这一段分析效率
The current benchmark design requires models to complete each task autonomously from a fixed specification without human intervention. Consequently, agents must decide in advance how much contextual reasoning, dependency checking, and tool preparation to incorporate. Since NeuroBench explicitly encourages broad consultation of available skills, enabling the skill library often leads to increased token consumption as models review more guidance, evaluate alternative pipelines, verify environment readiness, and consider whether the input data remains raw or partially processed.

This evaluation protocol may underestimate the effect of skill usage in real deployments. In neuroimaging workflows, explicit reasoning about missing dependencies, environment configuration, and data readiness is often necessary and aligned with user expectations. Under the current scoring rubric, however, such cautious reasoning can be partially penalized by the judge as extraneous to the narrow task requirement. The gains reported in Table~\ref{tab:neurobench-skillgain} should therefore be interpreted conservatively.

% 这一段分析有没有层级化的技能的差异
The normalized gain metric ($g$) provides a compact way to compare proportional improvements across models with different baseline strengths. In this evaluation, all models exhibit positive normalized gains, with MiniMax-M2.7 ($g = 0.1998$), Qwen 3 ($g = 0.1558$), and Claude-Sonnet-4.6 ($g = 0.1450$) showing the largest relative gains from skill integration. Runtime varies considerably across models, reflecting both inherent model latency and the additional overhead of skill-grounded reasoning. We plan to complement the current LLM-judge setup with multi-reviewer expert evaluations that better reflect user experience and scientific utility. Future extensions of NeuroBench will also include reproducibility stress tests on out-of-distribution data, direct executability assessment on live datasets, and evaluation of multi-agent collaboration patterns in neuroimaging workflows.

% NeuroBench 仍在建设，未来覆盖更多内容
We plan to complement the current LLM-based evaluation with multi-reviewer assessments conducted by domain experts, yielding indicators that better reflect real-user experience and scientific utility. Future extensions of NeuroBench will also incorporate reproducibility stress tests on out-of-distribution data and corrupted inputs, direct executability evaluations requiring actual tool invocation on real datasets, and systematic assessment of multi-agent collaboration patterns in neuroimaging workflows.

\section{Discussion}
% NeuroBench 对社区的作用
\subsection{Practical Value for the Community.} 
Beyond providing an agentic framework and benchmark, this release addresses a gap at the intersection of artificial intelligence and neuroimaging research. By offering a modular platform that can operate directly on raw data and standardized specifications, NeuroClaw enables neuroscience laboratories to move from fragile, irreproducible pipelines toward auditable, iterative experimental loops that more closely reflect scientific practice. Method developers can use NeuroBench to diagnose and improve agent behavior before large-scale deployment, while applied research groups can evaluate whether their existing agent stacks can sustain end-to-end data-to-model reasoning across diverse modalities and computing environments. Reviewers and editors also gain more transparent criteria for assessing methodological rigor. In addition, the released task specifications and harness engineering standards may serve as reusable templates for reproducible neuroimaging studies, lowering the barrier for interdisciplinary teams that combine clinical, imaging, and machine-learning expertise.

% 局限性
\subsection{Current Boundaries.} 
This work does not claim state-of-the-art performance or present exhaustive cross-system leaderboards. Instead, our focus is on executability, specification-driven validity, and loop-level reliability under realistic constraints. We do not evaluate model quality for clinical outcomes or make regulatory claims, as these questions require substantially larger-scale evaluation, prospective clinical trials, and domain-specific oversight beyond the scope of this study. We also acknowledge that the current NeuroBench tasks, while diverse, represent only a subset of real-world neuroimaging complexity. Expanding task difficulty, incorporating greater environmental stochasticity, and covering additional clinical populations remain important directions for future work.

% 伦理
\subsection{Ethical and Translational Considerations.} 
The automation of neuroimaging analysis carries both promise and responsibility. Because these systems interact with sensitive patient data and can influence downstream scientific conclusions and clinical decisions, questions of data governance, privacy preservation, bias amplification, and accountability arise even at the benchmark stage. We therefore view reproducibility and auditability as mechanisms for traceability, error diagnosis, and oversight. From a translational perspective, strong benchmark performance should be interpreted as infrastructure readiness rather than clinical readiness. Moving from reliable execution to safe deployment will require prospective validation, continuous expert supervision, characterization of failure modes, and task-specific governance frameworks. As agentic systems become more autonomous, the community will also need clearer standards for transparency, human oversight, and shared liability.

\section{Conclusion}
This paper presents NeuroClaw, a multi-agent system for neuroscience designed for executable and reproducible neuroimaging research, together with NeuroBench, a system-level benchmark with standardized specifications. NeuroClaw integrates raw-data-aware orchestration, a three-level skill hierarchy, and environment-aware runtime control to support an auditable discovery workflow that spans raw data processing, analysis, and iterative model-centric investigation without requiring preprocessed inputs or customized model code. NeuroBench evaluates these capabilities across major imaging modalities and representative datasets. By prioritizing specification clarity, shared task semantics, and adoption readiness, this work provides a foundation for community-wide evaluation and reproducible engineering progress in agentic neuroimaging research.

\clearpage

\section{References}
\begin{spacing}{0.9}
\bibliographystyle{naturemag}

\end{spacing}

\end{document}